\documentclass[12pt]{article}
\usepackage{graphicx}
\usepackage{amsmath,amssymb, dsfont, mathrsfs}
\usepackage{caption}
\usepackage{dcolumn}
\usepackage{bm}
\usepackage{float}
\usepackage{url}
\usepackage{subfig}
\usepackage{array}
\usepackage[table]{xcolor}
\usepackage{multicol}
\usepackage{multirow}
\usepackage{tabularx,ragged2e} 
\usepackage{booktabs}
\usepackage[authoryear, round]{natbib}

\newcommand{\Bs}[1]{\boldsymbol{#1}}

\newcommand{\bA}{\mathbf{A}}

\newcommand{\br}{\mathbf{r}}

\newcommand{\bs}{\mathbf{s}}

\newcommand{\bv}{\mathbf{v}}

\newcommand{\bw}{\mathbf{w}}

\newcommand{\bx}{\mathbf{x}}

\newcommand{\bsb}{\boldsymbol{b}}

\newcommand{\bsx}{\boldsymbol{x}}

\newcommand{\cT}{\mathcal{T}}

\newcommand{\bsalpha}{\boldsymbol{\alpha}}
\newcommand{\bsbeta}{\boldsymbol{\beta}}
\newcommand{\bsgamma}{\boldsymbol{\gamma}}
\newcommand{\bsGamma}{\boldsymbol{\Gamma}}
\newcommand{\bsomega}{\boldsymbol{\omega}}

\newcommand{\bstheta}{\boldsymbol{\theta}}

\newcommand{\bseta}{\boldsymbol{\eta}}

\newcommand{\bspsi}{\boldsymbol{\psi}}

\newcommand{\bsvPsi}{\boldsymbol{\varPsi}}

\newcommand{\bszeta}{\boldsymbol{\zeta}}
\newcommand{\bsxi}{\boldsymbol{\xi}}

\newcommand{\Pro}{\mathbb{P}}
\newcommand{\R}{\mathbb{R}}

 \usepackage[
                  breaklinks = true,
                 colorlinks = true,
                 linkcolor = blue,
                 urlcolor  = black, 
                 citecolor = blue,
                 anchorcolor = green,
                 ]{hyperref} 

\usepackage{empheq}

\begin{document}


\begin{center}
{\Large
	{\sc  Functional mixture-of-experts for classification}
}
\bigskip

Nhat Thien Pham and Fa\"icel Chamroukhi
\bigskip

{\small Normandie Univ, UNICAEN, CNRS, LMNO, 14000 Caen, France}

{\it
\href{mailto:nhat-thien.pham@unicaen.fr}{nhat-thien.pham@unicaen.fr}}, {\it \href{mailto:faicel.chamroukhi@unicaen.fr}{faicel.chamroukhi@unicaen.fr}}

\end{center}
\bigskip



{\bf Abstract.} We develop a mixtures-of-experts (ME) approach to the multiclass classification where the predictors are univariate functions. 
It consists of a ME model in which both the gating network and the experts network are constructed upon multinomial logistic activation functions with functional inputs. 
We perform a regularized maximum likelihood estimation in which the 
coefficient functions enjoy interpretable sparsity constraints on targeted derivatives. 
We develop an EM-Lasso like algorithm to compute the regularized MLE and evaluate the proposed approach on simulated and real data. 


\bigskip

\section{Introduction}
Introduced in \cite{jacobsME}, 
a mixtures of experts (ME) model 
can be defined 
as
\begin{equation}
    \label{eq: FME general}
\textstyle f(y|x) = \sum_{k=1}^K \text{Gate}_k(x)\,\text{Expert}_k(y|x),
\end{equation}
in which $f(y|x)$, the distribution of the response $y$ given the covariate $x$, is modeled as a mixture distribution with covariate-dependent mixing proportions $\text{Gate}_k(x)$, referred to as gating functions, and conditional mixture components $\text{Expert}_k(y|x)$, referred to as experts functions, $K$ being the number of experts.  
Some ME studies that may be mentioned here include ME for time series prediction \citep{ZeeviTS1996, Yml2003FinancialTS}, segmentation \citep{Chamroukhi-MRHLP-2013, chamroukhi_et_al_NN2009}, 
 ME for classification of gender and pose of human faces \citep{Gutta2000MixtureOE}, for social network data \citep{Gormley2010AMO}, among others.
For an overview of practical and theoretical aspects of ME modeling, the reader is referred to \cite{NguyenChamroukhi-MoE}. 
The study of ME for functional data analysis (FDA) \citep{RamsayAndSilvermanFDA2005}, is still however less investigated. In a recent study, we introduced in \cite{chamroukhi2022fME} a functional ME (FME) framework for regression and clustering of observed pairs of scalar responses and univariate functional  inputs.
 
\vspace{-.6cm}

\section{Functional Mixture-of-Experts for classification}
In this paper, we extend the FME framework for multiclass classification, derive adapted EM-like algorithms to obtain sparse and interpretable fit of the gating and experts network coefficients functions. 
Let $\{X_i(t), t \in \cT; Y_i\}_{i=1}^n$, be a  sample of $n$  i.i.d. data pairs where $Y_i\in \{1,\ldots,G\}$ is the class label of a functional predictor $X_{i}(\cdot)$, $G$ being the number of classes. 
In this case of functional inputs, a natural choice to model the conditional distribution $\text{Expert}_k(y|x) = \Pro(Y=y|X_{i}(\cdot))$ in \eqref{eq: FME general} is to use the functional multinomial logistic regression  modeling, see e.g., \cite{muller2005GFLM,James2002}, that is
{
\begin{eqnarray}
\label{eq: FME expert}
P(y_i|X_i(\cdot);\bsbeta_k) 
= \prod_{g=1}^G \left[\frac{\exp\left\{\beta_{kg,0} + \int_{\cT}X_i(t)\beta_{kg}(t)dt\right\}}
{1+\sum_{g^\prime=1}^{G-1}\exp\left\{\beta_{kg^\prime,0} + \int_{\cT}X_i(t)\beta_{kg^\prime}(t)dt\right\}}\right]^{y_{ig}},
\end{eqnarray}}whre 
$\bsbeta_k$ 
represents the set of coefficient functions $\{\beta_{kg}(t), t \in \cT\}$ and 
intercepts $\{\beta_{k,0}\}$
for $k\in[K]=\{1,\ldots,K\}$ and  $g\in[G]$, and $y_{ig}=\mathbb I_{\{y_i=g\}}$.
Similarly, a typical choice for the functional gating network $\text{Gate}_k(x) = \Pro(Z = k, X(\cdot))$ in \eqref{eq: FME general}, where $Z\in[K]$ is a hidden within-class clustering label, acting as weights for potential clusters $\{k\}$ in the heterogeneous functional inputs $X(\cdot)$
and which we denote as $\pi_{k}(X(\cdot))$, is to use a functional softmax function defined by
{\begin{eqnarray}\label{eq: FME gating}
\pi_{k}(X_i(\cdot);\bsalpha)
=\frac{\exp\{\alpha_{k,0} + \int_{\cT}X_i(t)\alpha_k(t)dt\}}{1+\sum_{k^\prime=1}^{K-1}\exp\{\alpha_{k^\prime,0} + \int_{\cT}X_i(t)\alpha_{k^\prime}(t)dt\}},
\end{eqnarray}}with 
$\bsalpha$ is composed of the set of coefficient functions $\{\alpha_{k}(t), t \in \cT\}$ and intercepts $\{\alpha_{k,0}\}$
for $k\in[K]$. 
Then, from \eqref{eq: FME expert} and \eqref{eq: FME gating} given $X_i(\cdot)$, the probability that $Y_i=y_i$, can be modeled by the following $K$-component FME model for classification
{
\begin{equation}\label{eq: FMEMR distribution}
P(y_i|X_i(\cdot);\bspsi) = \sum_{k=1}^{K} 
\pi_{k}(X_i(\cdot);\bsalpha)
P(y_i|X_i(\cdot);\bsbeta_k), \quad \bspsi=(\bsalpha,\bsbeta_1,\ldots,\bsbeta_K).
\end{equation}}
%

\vspace{-.5cm}
\subsection{Smooth functional representation}
In practice, $X_i(\cdot)$ is observed at a finite but large number of points on $\cT\subset \R$. 
In the perspective of parameter estimation, this results in estimating a very large number of coefficients $\beta$ and $\alpha$. 
In order to handle this high-dimensional problem, we consider a usual approach that projects the predictors and coefficient functions onto a family of reduced number of basis functions. 
Let $\bsb_r(t) = \left[b_1(t), \ldots, b_r(t)\right]^\top$ be a $r$-dimensional basis (B-spline, Wavelet, ...). Then, with $r,p,q\in\mathbb N$ sufficiently large, one can approximate $X_i(\cdot)$, $\alpha_k(\cdot)$ and $\beta_{kg}(\cdot)$ respectively by
{
\begin{equation}
X_i(t) = \bsx_i^\top \bsb_r(t), \quad
\alpha_k(t) = \bszeta_k^\top \bsb_p(t), \quad
\beta_{kg}(t) = \bseta_{kg}^\top \bsb_q(t).
\label{eq: projections}
\end{equation}}Here, $\bsx_i = ( x_{i1}, \ldots,  x_{ir})^\top$, with
$x_{ij} = \int_{\cT} X_i(t)b_j(t)dt$ for $j\in[r]$, is the vector of coefficients of $X_i(\cdot)$ in the basis $\bsb_r(t)$,
$\bszeta_k=(\zeta_{k,1},\ldots,\zeta_{k,p})^\top$, and $\bseta_{kg}=(\eta_{kg,1},\ldots,\eta_{kg,q})^\top$ are the  unknown coefficient vectors associated with the gating coefficient function $\alpha_k(\cdot)$ and
 the expert coefficient function $\beta_{kg}(\cdot)$ in the corresponding basis. In our case, we used B-spline bases.
Using the approximation of $X_i(\cdot)$ and $\alpha_k(\cdot)$ in \eqref{eq: projections}, the functional softmax gating network \eqref{eq: FME gating} can be represented by
{
\begin{eqnarray}\label{eq: FME gating projected}
\pi_{k}(\br_i;\bsxi)
=\frac{\exp\{\alpha_{k,0} + \br_i^\top \bszeta_k\}}
{1+\sum_{k^\prime=1}^{K-1}\exp\{\alpha_{k^\prime,0} + \br_i^\top \bszeta_{k^\prime} \}},
\end{eqnarray}}where $\br_i=\left[\int_{\cT}\bsb_r(t)\bsb_p(t)^\top dt\right]^\top \bsx_i$ is the design vector associated with the gating network and 
$\bsxi=((\alpha_{1,0},\bszeta_1^\top), \ldots, (\alpha_{K-1,0},\bszeta_{K-1}^\top))\allowbreak\in\R^{(p+1)(K-1)}$ is the unknown parameter vector of the gating network, to be estimated. 
In the same manner, using the approximations of $X_i(\cdot)$ and $\beta_{kg}(\cdot)$ in \eqref{eq: projections}, the expert conditional distribution \eqref{eq: FME expert} can be represented by
{\begin{equation}\label{eq: FME expert projected}
\textstyle P(y_i|\bx_i;\bstheta_k)
= \prod_{g=1}^G \left[\frac{\exp\left\{\beta_{kg,0} + \bx_i^\top \bseta_{kg} \right\}}
{1+\sum_{g^\prime=1}^{G-1}\exp\left\{\beta_{kg^\prime,0} + \bx_i^\top\bseta_{kg^\prime} \right\}}\right]^{y_{ig}},
\end{equation}}where
$\bx_i = \left[ \int_{\cT}\bsb_r(t)\bsb_q(t)^\top dt \right]^\top\bsx_i$ is the design vector associated with the expert network,
and $\bstheta_k=(\bstheta_{k1}^\top,\ldots,\bstheta_{k,G-1}^\top)^\top$, with $\bstheta_{kg}=(\beta_{kg,0},\bseta_{kg}^\top)^\top\in\R^{q+1}$ for $g\in[G-1]$, is the unknown parameter vector to be estimated of the expert distribution $k$.
Finally, combining \eqref{eq: FME gating projected} and \eqref{eq: FME expert projected}, the conditional distribution $P(y_i|X_i(\cdot);\bspsi)$ in \eqref{eq: FMEMR distribution} can be rewritten as
\[\label{eq: FMEMR distribution projected}
P(y_i|X_i(\cdot);\bsvPsi) = \sum_{k=1}^{K} 
\pi_{k}(\br_i;\bsxi)
P(y_i|\bx_i;\bstheta_k),
\]
where $\bsvPsi=(\bsxi^\top,\bstheta_1^\top,\ldots,\bstheta_K^\top)^\top$ is the unknown parameter vector of the model. 

\noindent {\it Parameter estimation:}
A maximum likelihood estimate (MLE) $\widehat{\bsvPsi}$ of $\bsvPsi$ can be obtained by using the EM algorithm for ME model for classification with vector data  
as in \cite{Chen1999}. We will refer to this approach as FME-EM.
To encourage sparsity in the model parameters $\bsvPsi$, one can perform penalized MLE by using the EM-Lasso algorithm as in \cite{prEMME-2019}. We refer to this approach as FME-EM-Lasso. 
%

\subsection{An interpretable sparse  
estimation of FME for classification}
Although fitting the FME model via EM-Lasso can accommodate sparsity in the parameters, it unfortunately does not ensure the reconstructed coefficient functions $\widehat \alpha_{k}(\cdot)$ and $\widehat \beta_{kg}(\cdot)$ are sparse and enjoy easy interpretable sparsity. 
To obtain interpretable and sparse fits for the coefficient functions, we simultaneously estimate the model parameters while constraining some targeted derivatives of the coefficient functions to be zero \citep{chamroukhi2022fME}. 
The construction of the interpretable FME model which we will fit with an adapted EM algorithm, is as follows.
 First, 
 in order to calculate the derivative of the  gating coefficient functions $\alpha_k(\cdot)$,
 let $\bA_p$ be the matrix of approximate $d_1$th and $d_2$th derivative of  $\bsb_p(t)$,  defined as in \cite{FLIRTI,chamroukhi2022fME} by
{\begin{eqnarray*}\label{eq:approx-derivatives-B(t)}
\bA_p &=& [\bA_p^{[d_1]} \bA_p^{[d_2]}]^{\top} = \left[D^{d_1} \bsb_p(t_1), \ldots, D^{d_1} \bsb_p(t_p), D^{d_2} \bsb_p(t_1), \ldots, D^{d_2} \bsb_p(t_p)\right]^{\top},
\end{eqnarray*}}where $D^d$ is the $d$th finite difference operator.  
Here $\bA_p^{[d_j]}$ is a square invertible matrix and $\bA_p\in\R^{2p\times p}$. Similarly,  to calculate the derivatives of the expert coefficient functions $\beta_{kg}(\cdot)$, 
let $\bA_q = [\bA_q^{[d_1]} \bA_q^{[d_2]}]^{\top} \in \R^{2q\times q}$ be the corresponding matrix defined for the $\bsb_q(t)$'s.
Now, if we define $\bsomega_{k} = \bA_p \Bs{\zeta}_{k}$ and denote $\bsomega_{k}= ({\bsomega^{[d_1]}_{k}}^\top, {\bsomega^{[d_2]}_{k}}^\top)^\top$, then $\bsomega^{[d_1]}_{k}$ and $\bsomega^{[d_2]}_{k}$ provide approximations to 
the $d_1$ and
the $d_2$ derivatives of the coefficient function $\alpha_{k}(\cdot)$, respectively, which we denote as $\alpha_{k}^{(d_1)}(\cdot)$ and $\alpha_{k}^{(d_2)}(\cdot)$.  Therefore, enforcing sparsity in $\bsomega_{k}$ will constrain $\alpha_{k}^{(d_1)}(\cdot)$ and $\alpha_{k}^{(d_2)}(\cdot)$ to be zero at most of time points.
Similarly, if we define $\bsgamma_{kg} = \bA_q \Bs{\zeta}_{k}$ and denote by $\bsgamma_{kg}= ({\bsgamma^{[d_1]}_{kg}}^\top, {\bsgamma^{[d_2]}_{kg}}^\top)^\top$, then we can derive the same regularization for the coefficient functions $\beta_{kg}(\cdot)$.
%
%
 From the definitions of $\bsomega_{k}$ and $\bsgamma_{kg}$ we can easily get the following  relations:
\begin{subequations}
  \begin{empheq}[left=\empheqlbrace]{alignat=2}
   \bszeta_{k} &= {\bA_p^{[d_1]}}^{-1} \bsomega_{k}^{[d_1]}
   \text{ and }
   \bsomega_{k}^{[d_2]}  = \bA_p^{[d_2]}{\bA_p^{[d_1]}}^{-1} \bsomega_{k}^{[d_1]}
 \label{iFME gating constrains}  \\
   \bseta_{kg} &= {\bA_q^{[d_1]}}^{-1} \bsgamma_{kg}^{[d_1]}
  \text{ and }
      \bsgamma_{kg}^{[d_2]}  = \bA_q^{[d_2]}{\bA_q^{[d_1]}}^{-1} \bsgamma_{kg}^{[d_1]}.
\label{iFME experts constrains}
  \end{empheq}
\end{subequations} 
Plugging the relation 
\eqref{iFME gating constrains}
into \eqref{eq: FME gating projected} one gets the following new representation for $\pi_{k}(\br_i;\bsxi)$
{
\begin{equation}\label{eq: iFME gating projected}
\textstyle \pi_{k}(\bs_i;\bw) 
= \frac{\exp{\{\alpha_{k,0}+ {\bs_i^\top \bsomega_k^{[d_1]}} \}}}
{1+\sum_{k^\prime=1}^{K-1}\exp{\{\alpha_{k^\prime,0}+{\bs_i^\top\bsomega^{[d_1]}_{k^\prime}} \}}},
\end{equation}}where $\bs_i = ({\bA^{[d_1]}_p}^{-1})^\top\br_i$ is now the new design vector and $\bw=(\alpha_{1,0},{\bsomega^{[d_1]}_1}^\top,\ldots,\alpha_{K-1,0},{\bsomega^{[d_1]}_{K-1}}^\top)^\top$, with $(\alpha_{K,0}, {\bsomega^{[d_1]}_{K}}^\top)^\top$ a null vector, is the unknown parameter vector of the gating network. Similarly, plugging 
\eqref{iFME experts constrains}
into \eqref{eq: FME expert projected} one obtains the new representation for $P(y_i|\bx_i;\bstheta_k)$:
{
\begin{equation}\label{eq: iFME expert projected}
\textstyle P(y_i|\bv_i;\bsGamma_k)
= \prod_{g=1}^G \left[\frac{\exp\left\{\beta_{kg,0} + {\bv_i^\top \bsgamma_{kg}^{[d_1]}}\right\}}
{1+\sum_{g^\prime=1}^{G-1}\exp\left\{\beta_{kg^\prime,0} + {\bv_i^\top\bsgamma_{k^\prime g}^{[d_1]}} \right\}}\right]^{y_{ig}},
\end{equation}}in which, $\bv_i = ({\bA^{[d_1]}_q}^{-1})^\top\bx_i$ is now  the new design vector and $\bsGamma_k = ( \beta_{kg,0}, {\bsgamma_{k^\prime g}^{[d_1]}}^\top )^\top$ is the unknown 
parameter vector of the expert network. Finally, gathering the gating network \eqref{eq: iFME gating projected} and the expert network \eqref{eq: iFME expert projected}, the iFME model for classification  is given by
$P(y_i|X_i(\cdot);\Bs{\Upsilon}) = \sum_{k=1}^{K} 
\pi_{k}(\bs_i;\bw)
P(y_i|\bv_i;\bsGamma_k),
$
where $\Bs{\Upsilon}=(\bw^\top,\bsGamma_1^\top,\ldots,\bsGamma_K^\top)^\top$ is the unknown parameter vector to be estimated. 
We perform penalized MLE by penalizing the ML via a Lasso penalization on the derivative coefficients $\bsomega_k$'s and $\bsgamma_{kg}$'s of the form Pen$_{\chi,\lambda}(\Bs{\Upsilon}) = \chi\sum_{k=1}^{K-1}\Vert \bsomega_k\Vert_1+ \lambda\sum_{k=1}^{K}\sum_{g=1}^{G-1}\Vert \bsgamma_{kg}\Vert_1$, with $\chi$ and $\lambda$ regularization constants. 
The estimation is performed by using an adaptation  to this classification context of the EM algorithm developed in \cite{chamroukhi2022fME}. The only difference resides in the maximization w.r.t. the  expert network parameters $\bsGamma_k$.


\vspace{-.6cm}

\section{Numerical results}
We conducted experiments by considering a $G=3$-class classification problem with a $K=2$-component FME model.
The simulation protocol will be detailed during the presentation due to lack of space here. 
The classification results obtained with the described algorithms FME-EM, FME-EM-Lasso and iFME-EM, as well as with functional multinomial logistic regression (FMLR), are given in Table \ref{Table: prediction performance} and show higher classification performance of the iFME-EM approach. 
{\small \begin{table}[htp!]
    \centering
    \def\arraystretch{.8}
    \begin{tabular}{r l l}
    \specialrule{1pt}{1pt}{1pt}
    Model & \multicolumn{2}{l}{Correct Classification Rate}\\
    \hline
    & Noise level: $\sigma^2_{\delta}=1$ & 
    Noise level: $\sigma^2_{\delta}=5$ \\
FME-EM & $.8560_{(.0199)}$ & $.8474_{(.0196)}$  \\ 
FME-EM-Lasso & $.9332_{(.0104)}$ & $.9178_{(.0142)}$  \\ 
iFME-EM & $\Bs{.9346_{(.0108)}}$ & $\Bs{.9219_{(.0127)}}$ \\ 
FMLR & $.7951_{(.0249)}$ & $.7922_{(.0270)}$ \\ 
	\specialrule{1pt}{2pt}{1pt}
	\end{tabular}
    \caption{Correct classification rates obtained on testing data. The reported values are averages on 100 samples with standard errors in parentheses.}\label{Table: prediction performance}
\end{table}}
\begin{figure}[h!]
\centering
\vspace{-0.1cm}
{\includegraphics[scale=0.7]{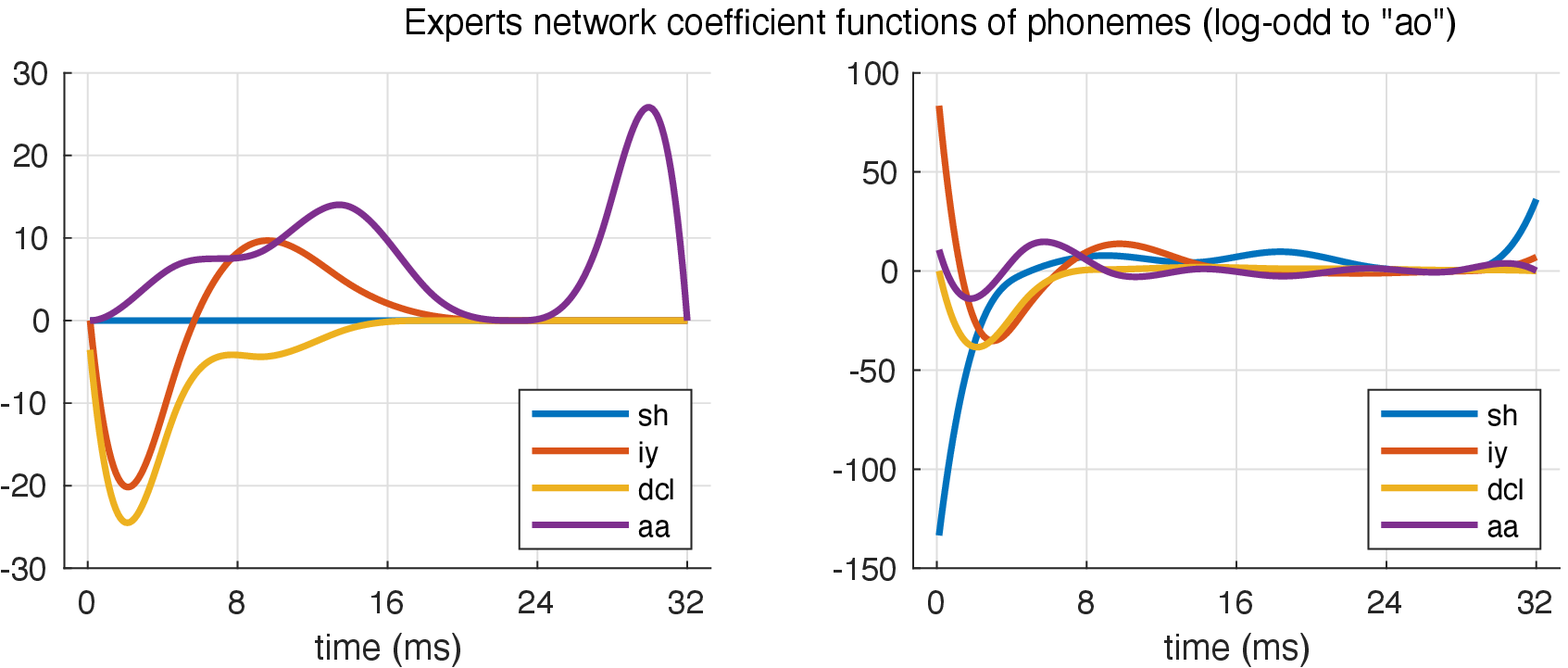}}\\
\vspace{0.2cm}
{\includegraphics[scale=0.7]{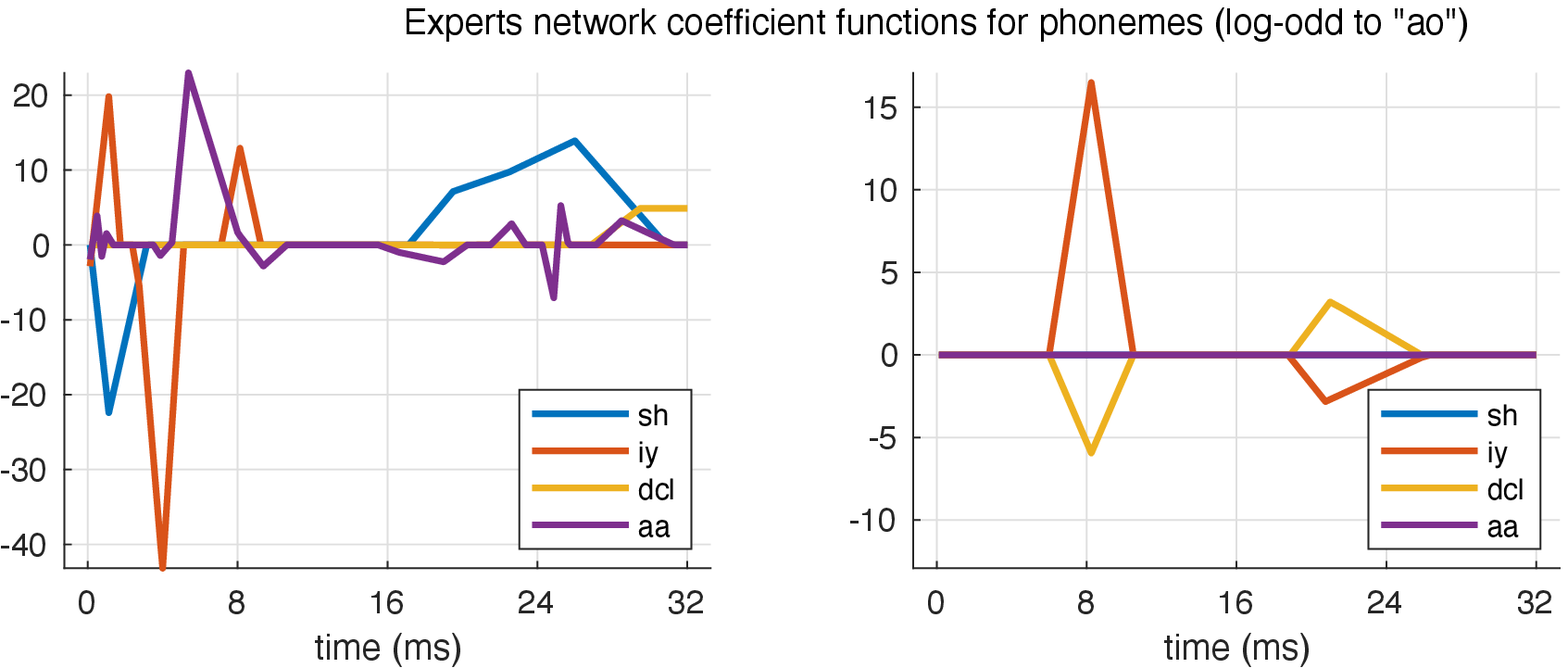}}
\caption{Results of (top) FME-EM-Lasso and (bottom) iFME-EM on phoneme data.}
\label{fig:phoneme results}
\end{figure}
We then applied the two algorithms allowing for sparsity (FME-EM-Lasso and iFME-EM)  to the well-known phoneme data \citep{Hastie95penalizeddiscriminant}. 
The data consists of $n=1000$ log-periodogram recordings of length $256$ each, used here as the univariate functional predictors, of five phonemes (the corresponding class labels).
The obtained averaged correct classification rate for the two approaches are more than 0.94 in mean. 
Figure \ref{fig:phoneme results} shows the estimated coefficient functions 
for the expert network  $\widehat \beta_{kg}(t)$ as functions of sampling time $t$, obtained by FME-EM-Lasso (top) and the iFME-EM (bottom); Here the iFME-EM is fitted with constraints on the zero and the second derivatives of the coefficients functions. The results show clearly sparse and piece-wise-linear gating and experts functions when using the iFME-EM approach.



\paragraph{Acknowledgement:} 
This research is supported by ANR SMILES ANR-18-CE40-0014.

\vspace{-.3cm}

\bibliographystyle{apalike}
\setlength{\bibsep}{0pt}
\bibliography{REFERENCES}

\begin{thebibliography}{}

\bibitem[Chamroukhi et~al., 2022]{chamroukhi2022fME}
Chamroukhi, F., Pham, T.~N., Hoang, V.~H., and McLachlan, G.~J. (2022).
\newblock Functional mixtures-of-experts.
\newblock {\em ArXiv preprint arXiv:2202.02249}.

\bibitem[Chamroukhi et~al., 2009]{chamroukhi_et_al_NN2009}
Chamroukhi, F., Sam\'{e}, A., Govaert, G., and Aknin, P. (2009).
\newblock Time series modeling by a regression approach based on a latent
  process.
\newblock {\em Neural Networks}, 22(5-6):593--602.

\bibitem[Chamroukhi et~al., 2013]{Chamroukhi-MRHLP-2013}
Chamroukhi, F., Trabelsi, D., Mohammed, S., Oukhellou, L., and Amirat, Y.
  (2013).
\newblock Joint segmentation of multivariate time series with hidden process
  regression for human activity recognition.
\newblock {\em Neurocomputing}, 120:633--644.

\bibitem[Chen et~al., 1999]{Chen1999}
Chen, K., Xu, L., and Chi, H. (1999).
\newblock Improved learning algorithms for mixture of experts in multiclass
  classification.
\newblock {\em Neural Networks}, 12(9):1229--1252.

\bibitem[Gormley and Murphy, 2010]{Gormley2010AMO}
Gormley, I.~C. and Murphy, T.~B. (2010).
\newblock A mixture of experts latent position cluster model for social network
  data.
\newblock {\em Statistical Methodology}, 7:385--405.

\bibitem[Gutta et~al., 2000]{Gutta2000MixtureOE}
Gutta, S., Huang, J.~R., Phillips, P.~J., and Wechsler, H. (2000).
\newblock Mixture of experts for classification of gender, ethnic origin, and
  pose of human faces.
\newblock {\em IEEE transactions on neural networks}, 11 4:948--60.

\bibitem[Hastie et~al., 1995]{Hastie95penalizeddiscriminant}
Hastie, T., Buja, A., and Tibshirani, R. (1995).
\newblock {Penalized Discriminant Analysis}.
\newblock {\em Annals of Statistics}, 23:73--102.

\bibitem[Huynh and Chamroukhi, 2019]{prEMME-2019}
Huynh, T. and Chamroukhi, F. (2019).
\newblock Estimation and feature selection in mixtures of generalized linear
  experts models.

\bibitem[Jacobs et~al., 1991]{jacobsME}
Jacobs, R.~A., Jordan, M.~I., Nowlan, S.~J., and Hinton, G.~E. (1991).
\newblock Adaptive mixtures of local experts.
\newblock {\em Neural Computation}, 3(1):79--87.

\bibitem[James, 2002]{James2002}
James, G.~M. (2002).
\newblock Generalized linear models with functional predictor variables.
\newblock {\em Journal of the Royal Statistical Society Series B}, 64:411--432.

\bibitem[James et~al., 2009]{FLIRTI}
James, G.~M., Wang, J., and Zhu, J. (2009).
\newblock {Functional linear regression that's interpretable}.
\newblock {\em Annals of Statistics}, 37(5A):2083--2108.

\bibitem[M{\"u}ller et~al., 2005]{muller2005GFLM}
M{\"u}ller, H.-G., Stadtm{\"u}ller, U., et~al. (2005).
\newblock Generalized functional linear models.
\newblock {\em Annals of Statistics}, 33(2):774--805.

\bibitem[Nguyen and Chamroukhi, 2018]{NguyenChamroukhi-MoE}
Nguyen, H.~D. and Chamroukhi, F. (2018).
\newblock Practical and theoretical aspects of mixture-of-experts modeling: An
  overview.
\newblock {\em Wiley Interdisciplinary Reviews: Data Mining and Knowledge
  Discovery}, pages e1246--n/a.

\bibitem[Ramsay and Silverman, 2005]{RamsayAndSilvermanFDA2005}
Ramsay, J.~O. and Silverman, B.~W. (2005).
\newblock {\em Functional Data Analysis}.
\newblock Springer Series in Statistics. Springer.

\bibitem[Y{\"u}ml{\"u} et~al., 2003]{Yml2003FinancialTS}
Y{\"u}ml{\"u}, M.~S., G{\"u}rgen, F.~S., and Okay, N. (2003).
\newblock Financial time series prediction using mixture of experts.
\newblock In {\em ISCIS}.

\bibitem[Zeevi et~al., 1996]{ZeeviTS1996}
Zeevi, A.~J., Meir, R., and Adler, R.~J. (1996).
\newblock Time series prediction using mixtures of experts.
\newblock In {\em Proceedings of NIPS'96}, page 309–315. MIT Press.

\end{thebibliography}
\end{document}